\documentclass[runningheads]{llncs}
\usepackage{graphicx}
\usepackage{cite}
\usepackage[hidelinks]{hyperref}
\usepackage{color}
\usepackage[flushleft]{threeparttable}
\usepackage{amssymb}
\usepackage{array}
\usepackage[misc]{ifsym}
\newcolumntype{x}[1]{>{\centering\arraybackslash\hspace{0pt}}p{#1}}

\makeatletter
\newcommand{\printfnsymbol}[1]{%
  \textsuperscript{\@fnsymbol{#1}}%
}

\makeatother
\begin{document}
% \title{Cross-Modality Domain Adaptation for Vestibular Schwannoma and Cochlea Segmentation}
\title{Unsupervised Domain Adaptation for Vestibular Schwannoma and Cochlea Segmentation via Semi-supervised Learning and Label Fusion}
\titlerunning{Unsupervised Domain Adaptation for VS and Cochlea Segmentation}
% If the paper title is too long for the running head, you can set
% an abbreviated paper title here

\author{Han Liu\inst{1}\thanks{equal contribution}\Letter \and
Yubo Fan\inst{1}\printfnsymbol{1} \and
Can Cui\inst{1} \and
Dingjie Su\inst{1} \and
Andrew McNeil\inst{2} \and
Benoit M. Dawant\inst{2}}

\authorrunning{H. Liu et al.}
% First names are abbreviated in the running head.
% If there are more than two authors, 'et al.' is used.
%
\institute{Department of Computer Science, Vanderbilt University, Nashville, TN, 37235\and 
Department of Electrical and Computer Engineering, Vanderbilt University, Nashville, TN, 37235\\
\email{han.liu@vanderbilt.edu}}
\maketitle              % typeset the header of the contribution
\begin{abstract}
Automatic methods to segment the vestibular schwannoma (VS) tumors and the cochlea from magnetic resonance imaging (MRI) are critical to VS treatment planning. Although supervised methods have achieved satisfactory performance in VS segmentation, they require full annotations by experts, which is laborious and time-consuming. In this work, we aim to tackle the VS and cochlea segmentation problem in an unsupervised domain adaptation setting. Our proposed method leverages both the image-level domain alignment to minimize the domain divergence and semi-supervised training to further boost the performance. Furthermore, we propose to fuse the labels predicted from multiple models via noisy label correction. In the MICCAI 2021 crossMoDA challenge\footnote{\url{https://crossmoda.grand-challenge.org/}}, our results on the final evaluation leaderboard showed that our proposed method has achieved promising segmentation performance with mean dice score of 79.9\% and 82.5\% and ASSD of 1.29 mm and 0.18 mm for VS tumor and cochlea, respectively. The cochlea ASSD achieved by our method has outperformed all other competing methods as well as the supervised nnU-Net.

\end{abstract}

\section{Introduction}
Vestibular schwannoma (VS) is a benign tumor that arises from the Schwann cells of the vestibular nerve, which connects the brain and the inner ear. To facilitate the follow-up and treatment planning of VS,  automatic methods to segment the VS tumors and the cochlea from magnetic resonance imaging (MRI) have been proposed \cite{1}. While the most commonly used modality for VS segmentation is contrast-enhanced T1 (ceT1), high-resolution T2 (hrT2) imaging has been demonstrated to be a possible alternative with less risk and lower cost\cite{2}. 

Supervised segmentation methods have shown to be effective for VS segmentation \cite{supervised}, but they require to fully annotate image data which may not be an option in practice. Weakly-supervised methods require less annotation efforts, such as scribbles and bounding boxes, and sometimes they even achieve a level of performance comparable to the supervised ones \cite{weakly}. In this work, we aim at segmenting the VS tumor and the cochlea in hrT2 without any hrT2 annotations during training. We consider the problem as an unsupervised domain adaptation (UDA) problem. Specifically, we are provided with a dataset consisting of ceT1 images and hrT2 images, but only the ceT1 images have the segmentation labels. 

There are mainly two types of methods to tackle the UDA problem, domain alignment and techniques based on semi-supervised learning (SSL). Domain alignment focuses on reducing the distribution discrepancy by optimizing some divergence metric \cite{31, 36} or via adversarial learning \cite{27, DANN}. Domain gaps can be bridged by image-level alignment \cite{a1, a2}, feature-level alignment \cite{b1, b2}, or the combination of the two \cite{c1}. On the other hand, due to the lack of labels in the target domain, techniques originating from SSL can be utilized to improve the performance. Zou et al. \cite{selftrain} and Zhang et al. \cite{e} use self-training based methods which iteratively generate pseudo labels and use them to retrain the network. To alleviate the negative impact from the noisy pseudo labels, learning from noisy labels has also received increasing interest. Motivated by \cite{cleanlab}, Zhang et al. \cite{f} use a confident learning module to characterize the label errors and correct them to achieve a more robust training. Mean teacher \cite{mt_original}, as another SSL-based technique, can be also used in UDA to provide competitive performance \cite{mt, g}. Inspired by previous works, we focus on exploring methods that combine image-level domain alignment and SSL for UDA.

\section{Methods}

\subsection{Problem Formulation}
For an unsupervised domain adaptation problem, we have access to a source domain $D^{S}=\{(x^{s}_{i}, y^{s}_{i})|i=1,2,\cdots,n_{s}\}$, and a target domain $D^{T}=\{x^{t}_{j}|j=1,2,\cdots,n_{t}\}$, where $Y^{S}$ and $Y^{T}$ share the same $K$ classes. In our case, source and target domains correspond to ceT1 and hrT2 respectively and $K=3$ representing background, VS and cochlea. We aim to train a segmentation network $F_{t}$ that learns from the source domain and is capable to achieve robust and accurate segmentation performance on the target domain, without accessing the target domain labels $Y^{T}$. 

\subsection{Image-level Domain Alignment}
Image-level domain alignment is a simple but effective method to tackle UDA problem by reducing the distribution mismatch at the image-level, i.e., pseudo image synthesis. Here, we propose to train the segmentation model $F_t$ with the pseudo target domain images $\tilde{X}^{T}$, which are generated by unpaired image-to-image translation. We explore both end-to-end training and two-stage training. For end-to-end training, we rely on the Contrastive Unpaired Translation (CUT) \cite{cut} as the backbone for image synthesis and add an extra segmentation module $F_t$ on top of the synthesized images. This method is referred as \textbf{CUTSeg}. We select CUT for unpaired image-to-image translation because it can be trained faster and is less memory-intensive than the CycleGAN \cite{cyclegan}, allowing more flexibility when adding the 3D CNN-based segmentation module. During training, we first train the CUT model alone till it achieves reasonable synthesis performance. Then we train the CUTSeg end-to-end with the CUT subnetwork initialized with the pre-trained weights and the segmentation module trained from scratch. For two-stage training, we use the CycleGAN  to generate pseudo hrT2 images $\tilde{X}^{T}$. To improve the data diversity, we train both 2D and 3D CycleGANs and collect pseudo images from different epochs. Lastly, we train a segmentation module $F_t$ using $\tilde{X}^{T}$.

\subsection{Semi-supervised Training}
Though image-level domain alignment can minimize the domain divergence, the unlabeled target domain images $X^{T}$ are not directly involved in training the segmentation model $F_{t}$. To overcome this limitation, we propose to adapt a semi-supervised learning method named Mean Teacher \cite{cv_mt} to make better use of $X^{T}$. Specifically, a student model along with a teacher model with the same network architecture are created and both models are initialized with the best model weights obtained from Section 2.2. In our semi-supervised setting, the labeled images are the pseudo hrT2 images while the unlabeled images are the real hrT2 images. During training, the labeled pseudo images are fed to the student model and the segmentation loss $L_{seg}$ is computed in a supervised manner. For unlabeled images, we first augment the same image twice with different intensity transformation parameters. The augmented images are then fed to the student model and the teacher model separately and a consistency loss $L_{con}$ is computed. As described in Equations 1-4, we use Dice loss \cite{v-net} and Cross-Entropy (CE) loss as $L_{seg}$ and Mean Squared Error (MSE) loss as $L_{con}$, where $p_{ik}$ is the predicted probability of the $i$th voxel at the $k$th output channel. Note that both $L_{seg}$ and $L_{con}$ are used to update the weights of the student model. The weights of the teacher model are updated as an exponential moving average (EMA) of the student weights, where the EMA decay coefficient is set as 0.99. As suggested in \cite{cv_mt}, the teacher prediction is more likely to be correct at the end of the training and thus the teacher model is taken as our final $F_{t}$.
\begin{equation}
    L_{seg} = L_{Dice} + L_{CE}
\end{equation}

\begin{equation}
L_{Dice} = 1- \frac{2\sum_{k}^{K}\sum_{i}^{N} p_{ik} y_{ik}}{\sum_{k}^{K}\sum_{i}^{N} p_{ik}^2 + \sum_{k}^{K}\sum_{i}^{N} y_{ik}^2}
\end{equation}

\begin{equation}
L_{CE}=-\frac{1}{N}\sum_{k}^{K}\sum_{i}^{N} y_{ik} \log p_{ik}
\end{equation}

\begin{equation}
    L_{con} = \frac{1}{N}\sum_{k}^{K}\sum_{i}^{N}(p_{ik}^{Teacher}-p_{ik}^{Student})^2
\end{equation}

\subsection{Noisy Label Correction as Label Fusion}
In this challenge, we have obtained three models (as shown in Figure 1) that were trained with different strategies and each model alone has achieved satisfactory result on the validation leaderboard. Specifically, the first model is obtained by two-stage training using the pseudo images from 2D CycleGAN, followed by a semi-supervised learning method, i.e., Mean Teacher. The second model is initialized with the teacher model and fine-tuned using the pseudo images from 3D CycleGAN. The third model is a CUTSeg model. Training details can be found in Section 3.1.

Empirically, ensembles tend to yield better predictive performance when there is a significant diversity among the models. Here, we propose to fuse the labels from different models by treating the label fusion task as a noisy label correction problem. We adapt a confident learning method called \textbf{CleanLab} \cite{cleanlab} which provides exact noise estimation and label error finding. Note that we use CleanLab to directly fuse labels at the inference phase rather than update the pseudo labels iteratively during training. Specifically, we first obtain the softmax outputs of two models and convert one output to a one-hot encoded label mask. The one-hot encoded mask is considered as a 'noisy label' and corrected by the softmax outputs from the other model. Once the labels from the first two models are fused, the fused labels are treated as noisy labels and fused again with the softmax outputs from the remaining model. The labels fused from the three models are used as our final predictions.
\begin{figure}[h]
\includegraphics[width=1\columnwidth]{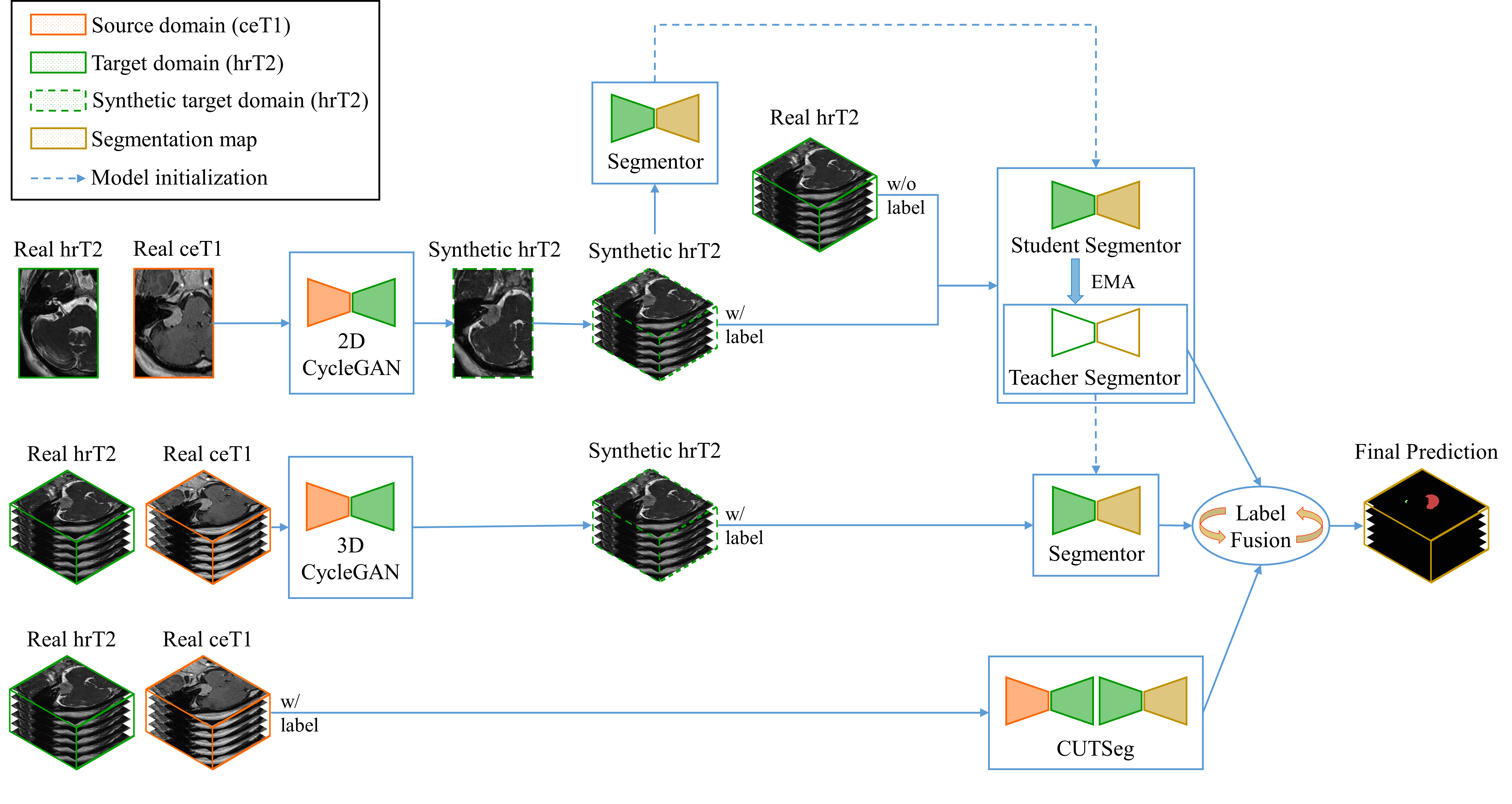}
\centering
\caption{The schematic diagram of our proposed method} \label{fig1}
\end{figure} 

\section{Experiments and Results}
\subsection{Experimental Design}
In this section, we describe different methods in our experiments and summarize the training details in Table 1. 

\subsubsection{Methods \#1-3} First, we compare different segmentation backbones including 2D nnU-Net, 3D nnU-Net, and the 2.5D U-Net proposed in \cite{supervised}, which are referred to as method \#1-3. \subsubsection{Method \#4} We utilize a semi-supervised learning method named Mean Teacher to leverage the unlabeled real T2 images. Details are described in Section 2.3. \subsubsection{Method \#5} Here, we explore the feasibility of using self-training to improve the segmentation performance of Mean Teacher. Specifically, we use the mean teacher model to obtain the pseudo labels on the real T2 images. Then we fine-tune the teacher model obtained from method \#4 with the pseudo-labeled real T2 images. The pseudo labels are iteratively updated at the end of each epoch by a confident learning method \cite{cleanlab}. Note that methods \#1-4 are trained using pseudo images generated from 2D CycleGAN. \subsubsection{Method \#6} We incorporate more training data to further boost the performance. Specifically, we fine-tune the teacher model from method \#4 on additional pseudo images generated from 3D CycleGAN. \subsubsection{Method \#7} Here, we train an end-to-end CUTSeg model and the details can be found in Section 2.2. \subsubsection{Methods \#8-9} Lastly, we fuse the predictions from different models by CleanLab as described in section 2.4.

\begin{table}[h]
% \centering
\caption{Experimental design}
\begin{tabular}{cx{3.5cm}cx{3.5cm}cx{3.5cm}cx{3.5cm}cx{3.5cm}}%{c{0.8cm}c{2.5cm}c{2.5cm}c{2.5cm}c{5cm}}%{ccccc}
\hline
\# & Method              & Training image   & Training label        & Model init. \\ \hline
1  & nnU-Net (2D)        & S-T2 (2D)        & T1                    & scratch              \\
2  & nnU-Net (3D)        & S-T2 (2D)        & T1                    & scratch              \\
3  & U-Net (2.5D)           & S-T2 (2D)        & T1                    & scratch              \\
4  & Mean teacher        & S-T2 (2D) + R-T2 & T1                    & \#3                  \\
5  & \#4 + self-training & R-T2             & \#4 (pseudo label) & \#4 (teacher)                 \\
6  & \#4 + fine-tuning    & S-T2 (3D)        & T1                    & \#4 (teacher)                  \\
7  & CUTSeg              & R-T1 + R-T2      & T1                    & scratch              \\
8  & \#4 $\circ$ \#6               & -                & -                     & -                    \\
9  & \#4 $\circ$ \#6 $\circ$ \#7           & -                & -                     & -                    \\ \hline
\end{tabular}
\begin{tablenotes}
      \small
      \item In the method column, $\circ$ represents label fusion using CleanLab. In the training data column, S and R represent synthetic data from CycleGAN and real data, respectively. 2D and 3D represent synthetic data generated from 2D CycleGAN and 3D CycleGAN, respectively.
\end{tablenotes}
\end{table}

\subsection{Data and Implementation}
The dataset was released by the MICCAI challenge crossMoDA 2021 \cite{summary}. All images were obtained on a 32-channel Siemens Avanto 1.5T scanner using a Siemens single-channel head coil \cite{data}. ceT1 images have in-plane resolution of 0.41×0.41 mm and slice thickness of 1.0 or 1.5 mm. For hrT2 images, the in-plane resolution varies from 0.47×0.47 mm to 0.55×0.55 mm and slice thickness is 1.0 or 1.5 mm. The VS and cochleae were manually segmented in consensus by the treating neurosurgeon and physicist using both the ceT1 and hrT2 images. We randomly split the images into 185 and 25 for training and validation respectively. Since the Field of View (FoV) of the source and target domain images varies significantly, we crop each image into a cubic box, or ROI, using single-atlas registration \cite{ants}. As shown in Figure 2, the ROI on the atlas image is manually cropped around the right side of the brain. To obtain the ROI on the left side, we flip the volume left-to-right before performing registration. 

For preprocessing, in two-stage training (methods \#1-6), we resample the images to the most common spacing in the target domain, i.e., (0.46875, 0.468975, 1.5) and normalize the intensity to [0, 1]. In end-to-end training (method \#7), we first train a CUTSeg model for 82 epochs with an auxiliary consistency loss, which is a Mean Absolute Error (MAE) loss between the segmentation result of the real hrT2 image and the prediction from the two-stage training model. This model is used to make inferences for the testing images with in-plane resolution less than 0.5 mm. We use another CUTSeg model which is fine-tuned on the hrT2 images with in-plane resolution more than 0.5 mm to make inference for the testing images with such resolution. For all our segmentors, we use the 2.5D U-Net architecture proposed in \cite{supervised}. For post-processing, we first reduce the false positive VS prediction by removing the isolated components whose center is more than 15 voxels along the z-axis from the adjacent cochlea center. Then we take the largest connected components for both VS and cochlea within each ROI.

\begin{figure}[h]
\includegraphics[width=1\columnwidth]{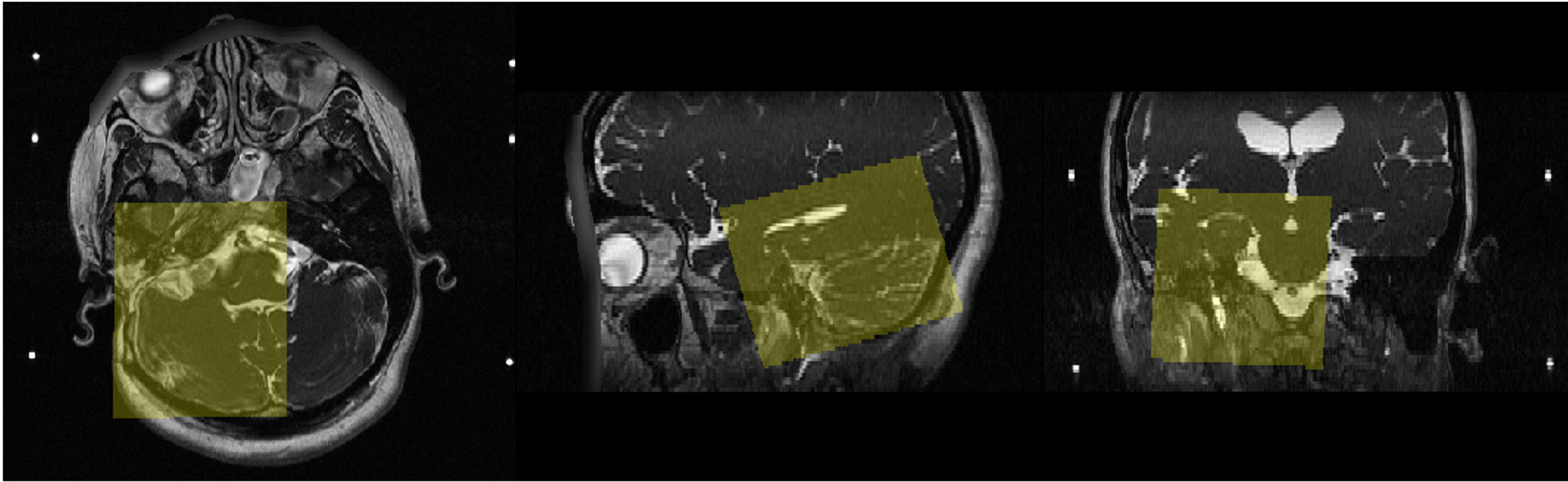}
\centering
\caption{An illustration of our cropped ROI on the target domain atlas image.} \label{fig2}
\end{figure} 

For training, we use Adam optimizer with weight decay $10^{-4}$ and batch size $1$. The learning rates are initialized to $5\times10^{-4}$, $5\times10^{-5}$ and $2\times10^{-4}$ for two-stage training, Mean Teacher, and CUTSeg, respectively. The hyperparameters are determined by grid-search within the range of $10^{-2}$ to $10^{-6}$. The best hyperparameters are selected based on the segmentation performance on our own validation set. The CNNs are implemented in PyTorch \cite{pytorch} and MONAI on a Ubuntu desktop with an NVIDIA RTX 2080 Ti GPU. For quantitative evaluation, we measure the Dice score and Average Symmetric Surface Distance (ASSD) between the segmentation results and the ground truth.

\begin{table}[h]
\centering
\caption{Quantitative results on validation leaderboard}
\begin{tabular}{ccccccc}
\hline
   &               &         \multicolumn{2}{c}{Dice$\uparrow$ (\%)}             & \multicolumn{2}{c}{ASSD$\downarrow$ (mm)}           \\ \cline{3-6} 
\# & Method            & VS                  & Cochlea             & VS                 & Cochlea            \\ \hline
1  & nnU-Net (2D)   & 72.90±22.77         & 68.14±12.62         & 1.32±2.83          & 1.27±3.65          \\
2  & nnU-Net (3D)      & 71.77±27.29         & 79.95±4.12          & 2.71±5.09          & 0.20±0.07          \\
3  & U-Net (2.5D)        & 74.81±22.14         & 80.39±3.18          & 1.37±2.63          & 0.22±0.05          \\
4  & Mean teacher   & 80.81±9.09          & 80.50±6.36          & 0.63±0.30          & 0.20±0.06          \\
5  & \#4 + self-training       & 80.73±5.21          & 81.16±4.25          & 0.69±0.35          & 0.19±0.04          \\
6  & \#4 + fine-tuning          & 81.66±10.5          & 81.24±3.52          & 0.61±0.37          & 0.20±0.06          \\
7  & CUTSeg             & 81.00±7.50          & 81.45±3.53          & 0.66±0.29          & 0.19±0.06          \\
8  & \#4 $\circ$ \#6                       & 82.08±4.77          & 81.46±3.51          & 0.58±0.31          & 0.19±0.06          \\
9  & \#4 $\circ$ \#6 $\circ$ \#7                      & \textbf{83.02±7.72} & \textbf{82.20±3.10} & \textbf{0.57±0.27} & \textbf{0.18±0.05} \\ \hline
\end{tabular}
\end{table}

\subsection{Experimental Results}
Table 2 shows the evaluation metrics on the validation leaderboard of our developed methods. By comparing method \#1-3, we notice that the 2.5D U-Net architecture inspired by \cite{supervised} outperforms the 2D and 3D nnU-Nets and thus is used as the segmentation backbone in all our experiments. We also find that by incorporating the real unlabeled T2 images, our Mean Teacher model (method \#4) is able to increase the VS dice score from 74.81\% to 80.81\%, demonstrating the effectiveness of our semi-supervised learning strategy. We observe that a smaller learning rate and appropriate model initialization, i.e., the pre-trained segmentation model weights from method \#3, are critical for the effectiveness of MT. Since the Mean Teacher achieved the best performance among method \#1-4 on leaderboard during validation phase, we use the best weights obtained by method \#4 for model initialization for other methods, i.e., methods \#5-6. By comparing method \#4 and \#5, we observe that self-training slightly outperforms the mean teacher model on cochlea but underperforms on VS tumor.  By fine-tuning with more training data, method \#6 shows slight improvements on both VS and cochlea compared to method \#4.  Furthermore, we find that the CUTSeg model (method \#7) achieves comparable segmentation performance to method \#4. Lastly, in method \#8 and \#9, we show that though CleanLab is unable to improve performance by iteratively updating labels during training (method \#5), it is however an effective method to fuse the predictions from different models during the inference phase. In the evaluation phase, our proposed method (method \#9) achieved dice scores of 79.9\% and 82.5\% and ASSD of 1.29 mm and 0.18 mm for VS tumor and cochlea, respectively. We note that the ASSD of cochlea achieved by our method (0.18 mm) is the lowest among all unsupervised methods and even lower than that achieved by the supervised nnU-Net (0.22 mm). 

\begin{figure}[h]
\includegraphics[width=1\columnwidth]{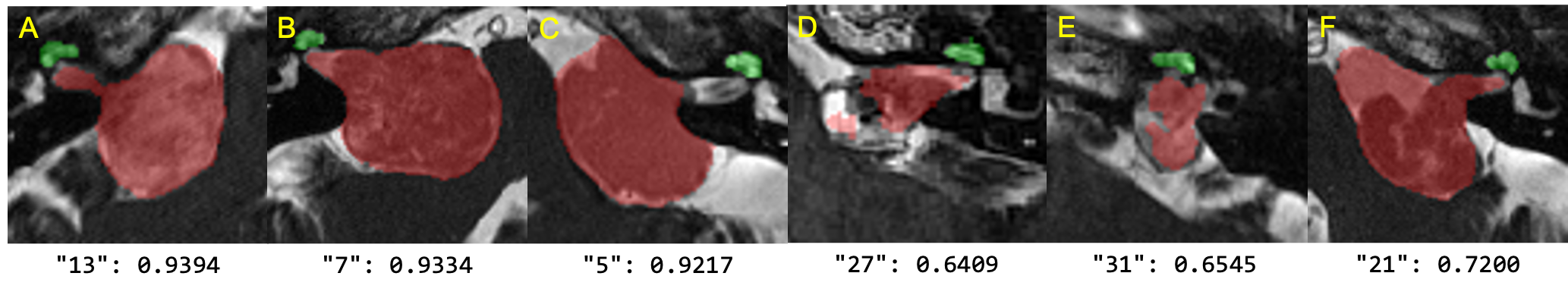}
\centering
\caption{Qualitative results on the validation set. A to C and D to F display the best and worst VS segmentation results. The image ID and the corresponding dice score are also shown. Our experiments show that VS tumors with higher inhomogeneity, e.g., D, are more difficult to segment.} \label{fig3}
\end{figure} 

\section{Discussion}
\subsection{ROI Selection Strategy}
In this challenge, we observe that the FoV of source and target domain images varies significantly. To minimize the impact of FoV difference, we manually determine a cubic box as ROI and use rigid registration to obtain the ROI from each volume as described in section 3.1. Note that our ROI covers either the left or the right side region around the cochlea while most other teams use the ROIs that cover both cochleae. Arguably, our ROI selection strategy has both a positive and a negative impact on the downstream segmentation tasks. First, this strategy helps to increase the amount of training data since two ROIs can be extracted from one volume. Second, by flipping the right ROI to the left, the orientation and relative position of the cochlea within each ROI remain almost the same, which is greatly beneficial for the accuracy and robustness of cochlea segmentation. Based on the results from the validation leaderboard, all our attempted methods have achieved around 80\% dice scores, which are very competitive against other teams' results. Moreover, in the evaluation leaderboard, our method has achieved an ASSD of 0.18 mm which is lower than the supervised nnU-Net of 0.22 mm, suggesting the benefits of our ROI selection strategy on cochlea segmentation. However, this strategy inevitably introduces challenges for VS segmentation. Because VS tumor can be either fully/partially included in both left and right ROIs, the merging scheme for combining the VS segmentation results from both ROIs needs to be carefully designed. In the future, it may be interesting to explore whether using ROIs with different FoVs for VS tumor and cochlea can help improve the results. 

\subsection{Self-training Strategy}
Our experiments show that our self-training strategy cannot help to improve the segmentation performance. In contrast, both the top-two teams on the leaderboard found the self-training strategy very beneficial for their performance. Specifically, the 1st-place team updated the noisy labels by manually inspecting their qualities and filtered out the unreliable labels. On the other hand, the 2nd-place team considered the voxels with the higher probabilities as confident labels to train the next epoch. Hence, we speculate that the effectiveness of self-training may heavily depend on the filtering strategy for noisy pseudo labels. Besides, in our self-training experiments, we only used the real unlabeled data while the aforementioned teams used the combined data, i.e., pseudo T2 and real T2 images. The combined data might also be a key factor for the effectiveness of self-training. In the future, an exciting direction to improve the performance would be to develop an automatic and smart filtering strategy for self-training.

\section{Conclusion}
In this work, we exploited the image-level domain alignment and semi-supervised training to tackle the unsupervised domain adaptation segmentation problem. The results on the validation and evaluation leaderboard of crossMoDA challenge show that our proposed method can yield promising segmentation performance on VS tumor and cochlea on hrT2 MRI images without labels. 

% ---- Bibliography ----
%
% BibTeX users should specify bibliography style 'splncs04'.
% References will then be sorted and formatted in the correct style.
%
% \bibliographystyle{splncs04}
% \bibliography{mybibliography}

\begin{thebibliography}{8}

\bibitem{1}
Vokurka, Elizabeth A., et al. "Using Bayesian tissue classification to improve the accuracy of vestibular schwannoma volume and growth measurement." American journal of neuroradiology 23.3 (2002): 459-467.

\bibitem{2}
Coelho, Daniel H., et al. "MRI surveillance of vestibular schwannomas without contrast enhancement: clinical and economic evaluation." The Laryngoscope 128.1 (2018): 202-209.

\bibitem{supervised}
Wang, Guotai, et al. "Automatic segmentation of vestibular schwannoma from T2-weighted MRI by deep spatial attention with hardness-weighted loss." International Conference on Medical Image Computing and Computer-Assisted Intervention. Springer, Cham, 2019.

\bibitem{weakly}
Dorent, Reuben, et al. "Scribble-based Domain Adaptation via Co-segmentation." International Conference on Medical Image Computing and Computer-Assisted Intervention. Springer, Cham, 2020.

\bibitem{31}
Lee, Chen-Yu, et al. "Sliced wasserstein discrepancy for unsupervised domain adaptation." Proceedings of the IEEE/CVF Conference on Computer Vision and Pattern Recognition. 2019.

\bibitem{36}
Long, Mingsheng, et al. "Learning transferable features with deep adaptation networks." International conference on machine learning. PMLR, 2015.

\bibitem{27}
Hoffman, Judy, et al. "Cycada: Cycle-consistent adversarial domain adaptation." International conference on machine learning. PMLR, 2018.

\bibitem{DANN}
Ganin, Yaroslav, et al. "Domain-adversarial training of neural networks." The journal of machine learning research 17.1 (2016): 2096-2030.

\bibitem{selftrain}
Zou, Yang, et al. "Unsupervised domain adaptation for semantic segmentation via class-balanced self-training." Proceedings of the European conference on computer vision (ECCV). 2018.

\bibitem{mt}
Perone, Christian S., et al. "Unsupervised domain adaptation for medical imaging segmentation with self-ensembling." NeuroImage 194 (2019): 1-11.

\bibitem{cut}
Park, Taesung, et al. "Contrastive learning for unpaired image-to-image translation." European Conference on Computer Vision. Springer, Cham, 2020.

\bibitem{cyclegan}
Zhu, Jun-Yan, et al. "Unpaired image-to-image translation using cycle-consistent adversarial networks." Proceedings of the IEEE international conference on computer vision. 2017.

\bibitem{cv_mt}
Tarvainen, Antti, and Harri Valpola. "Mean teachers are better role models: Weight-averaged consistency targets improve semi-supervised deep learning results." arXiv preprint arXiv:1703.01780 (2017).

\bibitem{v-net}
Milletari, F., Navab, N., \& Ahmadi, S. A. (2016, October). V-net: Fully convolutional neural networks for volumetric medical image segmentation. In 2016 fourth international conference on 3D vision (3DV) (pp. 565-571). IEEE.

\bibitem{cleanlab}
Northcutt, Curtis, Lu Jiang, and Isaac Chuang. "Confident learning: Estimating uncertainty in dataset labels." Journal of Artificial Intelligence Research 70 (2021): 1373-1411.

\bibitem{summary}
Dorent, R., Kujawa, A., Ivory, M., Bakas, S., Rieke, N., Joutard, S., ... \& Vercauteren, T. (2022). CrossMoDA 2021 challenge: Benchmark of Cross-Modality Domain Adaptation techniques for Vestibular Schwnannoma and Cochlea Segmentation. arXiv preprint arXiv:2201.02831.

\bibitem{data}
Shapey, J., Kujawa, A., Dorent, R., Wang, G., Dimitriadis, A., Grishchuk, D., Pad-dick, I., Kitchen, N., Bradford, R., Saeed, S.R., Bisdas, S., Ourselin, S., Vercauteren,T.: Segmentation of vestibular schwannoma from mri — an open annotated datasetand  baseline  algorithm.  Scientific  Data  (2021),  in  press.  Preprint  available  athttps://doi.org/10.1101/2021.08.04.21261588medRXiv:10.1101/2021.08.04.21261588

\bibitem{ants}
Avants, Brian B., et al. "A reproducible evaluation of ANTs similarity metric performance in brain image registration." Neuroimage 54.3 (2011): 2033-2044.

\bibitem{pytorch}
Paszke, Adam, et al. "Pytorch: An imperative style, high-performance deep learning library." Advances in neural information processing systems 32 (2019): 8026-8037.

\bibitem{a1}
Zhang, Yue, et al. "Task driven generative modeling for unsupervised domain adaptation: Application to x-ray image segmentation." International Conference on Medical Image Computing and Computer-Assisted Intervention. Springer, Cham, 2018.

\bibitem{a2}
Bousmalis, Konstantinos, et al. "Unsupervised pixel-level domain adaptation with generative adversarial networks." Proceedings of the IEEE conference on computer vision and pattern recognition. 2017.

\bibitem{b1}
Zhang, Yifan, et al. "Collaborative unsupervised domain adaptation for medical image diagnosis." IEEE Transactions on Image Processing 29 (2020): 7834-7844.

\bibitem{b2}
Chang, Wei-Lun, et al. "All about structure: Adapting structural information across domains for boosting semantic segmentation." Proceedings of the IEEE/CVF Conference on Computer Vision and Pattern Recognition. 2019.

\bibitem{c1}
Chen, Cheng, et al. "Unsupervised bidirectional cross-modality adaptation via deeply synergistic image and feature alignment for medical image segmentation." IEEE transactions on medical imaging 39.7 (2020): 2494-2505.

\bibitem{d}
Zhang, Pan, et al. "Prototypical pseudo label denoising and target structure learning for domain adaptive semantic segmentation." Proceedings of the IEEE/CVF Conference on Computer Vision and Pattern Recognition. 2021.

\bibitem{e}
Zhang, Qiming, et al. "Category Anchor-Guided Unsupervised Domain Adaptation for Semantic Segmentation." Advances in Neural Information Processing Systems 32 (2019): 435-445.

\bibitem{f}
Zhang, Minqing, et al. "Characterizing label errors: Confident learning for noisy-labeled image segmentation." International Conference on Medical Image Computing and Computer-Assisted Intervention. Springer, Cham, 2020.

\bibitem{g}
Zhao, Ziyuan, et al. "MT-UDA: Towards Unsupervised Cross-modality Medical Image Segmentation with Limited Source Labels." International Conference on Medical Image Computing and Computer-Assisted Intervention. Springer, Cham, 2021.

\bibitem{mt_original}
Tarvainen, Antti, and Harri Valpola. "Mean teachers are better role models: Weight-averaged consistency targets improve semi-supervised deep learning results." Advances in Neural Information Processing Systems 30 (2017).


\end{thebibliography}
%

\end{document}